\documentclass[11pt,a4paper]{article}

\usepackage[utf8]{inputenc}
\usepackage[T1]{fontenc}
\usepackage[english]{babel}
\usepackage{amsmath,amssymb}
\usepackage{hyperref}
\usepackage{url}
\usepackage{graphicx}
\usepackage{microtype}
\usepackage{textcomp}
\usepackage{booktabs}
\usepackage[margin=2.5cm]{geometry}
\usepackage{needspace}
\usepackage{enumitem}
\setlist[itemize]{topsep=4pt, partopsep=0pt, itemsep=2pt, parsep=0pt, leftmargin=1.8em}
\setlist[enumerate]{topsep=4pt, partopsep=0pt, itemsep=2pt, parsep=0pt, leftmargin=2.2em}
\usepackage{xcolor}
\usepackage{fancyvrb}
\RecustomVerbatimEnvironment{verbatim}{Verbatim}{%
  fontsize=\small, xleftmargin=1.5em, framesep=8pt,
  frame=leftline, framerule=0.6pt, rulecolor=\color{black!30}}
\usepackage{tikz}
\usetikzlibrary{arrows.meta,positioning}
\tikzset{
  nd/.style={draw, rounded corners=2pt, align=center, inner xsep=5pt,
             inner ysep=4pt, minimum height=8mm, font=\small},
  cdn/.style={nd, font=\small\ttfamily},
  fl/.style={-{Stealth[length=2.2mm]}, semithick},
  bi/.style={{Stealth[length=2.2mm]}-{Stealth[length=2.2mm]}, semithick}
}
\hypersetup{
    pdftitle={Benchy: towards a universal language for task-oriented AI benchmarks},
    pdfauthor={Francis F Daniel, Mauro Ibañez, Francis Perelman, Marian Basti},
    pdfkeywords={AI evaluation, benchmarks, semantic language, intermediate representation, task-oriented AI},
    colorlinks=true,
    linkcolor=blue,
    citecolor=blue,
    urlcolor=blue
}

\newcommand{\gloss}[1]{\begin{quote}\itshape #1\end{quote}}

\title{Benchy: towards a universal language for task-oriented AI benchmarks}
\author{
  Francis F Daniel\textsuperscript{1} \and
  Mauro Ibañez\textsuperscript{1} \and
  Francis Perelman\textsuperscript{2} \and
  Marian Basti\textsuperscript{1}
}
\date{September 23, 2026}

\begin{document}
\maketitle

\begin{center}
\textsuperscript{1}SURUS \quad \textsuperscript{2}Independent\\[0.3em]
Contact: \texttt{francis@surus.lat}
\end{center}

\begin{abstract}
Benchy is a semantic language and execution engine for benchmarking AI programs. A benchmark is completely specified by a program, a scoring function, and a dataset, $B=(P,S,D)$, and is separate from the AI-system taking it; a run binds the two, $R=(B,\mathrm{AI})$. Benchmarks are authored as canonical YAML in which each semantic concept has one valid syntax, classified by a shared task/domain/language ontology, and deterministically compiled into a canonical JSON intermediate representation that the engine executes. Compilation changes representation, not meaning: it does not repair invalid definitions or inject hidden defaults. Programs use fixed schemas of named input and output fields, the leaf output fields are the scoring dimensions, and the engine exposes one universal runtime contract --- a named-field input object in, a named-field output object out --- to which external AI-systems adapt at the boundary, so integration mechanics never propagate into benchmark semantics. This paper gives the semantic object model, the ontology and task-to-program validation rule, the scoring and failure semantics, the compilation and execution architecture, and the scope of the current language. An appendix fixes the normative engineering contract for the first engine implementation.
\end{abstract}


\tableofcontents
\newpage

\textit{SURUS Team \textperiodcentered{} Buenos Aires \textperiodcentered{} 2026}

\section{Overview}

Benchy is a semantic language and execution engine for benchmarking AI programs.

A \textbf{program} defines a typed input/output contract. An \textbf{AI-system} is any AI-based implementation of that program: a single AI model; an AI-node such as a model with an optimized prompt and fixed task behavior; a composition of AI models or AI programs; AI components intermingled with explicit or deterministic code; or an agent/workflow composed from those elements.

A benchmark is separate from the AI-system taking it.

\begin{equation}
\boxed{B=(P,S,D)}
\end{equation}
\gloss{Benchmark $B$ is completely specified by program $P$, scoring function $S$, and dataset $D$.}

A benchmark run binds that benchmark to an AI-system:

\begin{equation}
\boxed{R=(B,\mathrm{AI})}
\end{equation}
\gloss{Run $R$ evaluates AI-system $\mathrm{AI}$ on benchmark $B$.}

The design follows a small number of decisions:

\begin{itemize}
  \item YAML is the canonical semantic definition authored by humans, agents, and the UI.
  \item Each semantic concept has one valid YAML syntax.
  \item Valid YAML is deterministically compiled into a canonical JSON intermediate representation (IR).
  \item Compilation changes representation, not meaning. It does not repair invalid definitions or inject hidden defaults.
  \item Types express semantic constraints, not storage representation.
  \item Programs use fixed schemas composed of named input and output fields.
  \item Variable-length output collections are outside the current program model.
  \item The output schema determines the fixed scoring dimensions.
  \item Task, domain, and language come from one shared SURUS ontology registry.
  \item Benchy defines one universal runtime contract from the declared program schema.
  \item External AI-systems adapt to that contract at the boundary.
  \item Integration-specific mechanics do not propagate into benchmark semantics or the engine.
\end{itemize}

The authoring and execution path is:

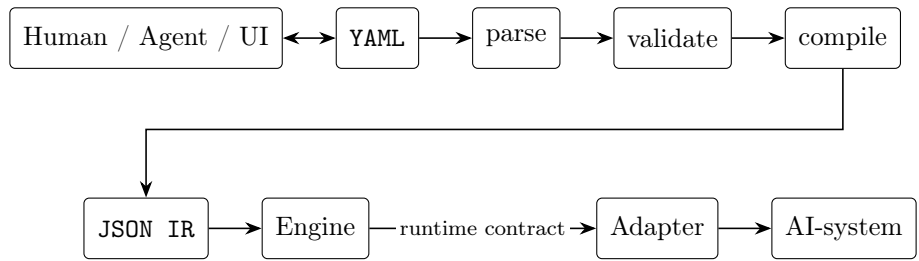
\begin{center}
\begin{tikzpicture}[node distance=6mm and 7mm]
  \node[nd]                     (hu)    {Human / Agent / UI};
  \node[cdn, right=of hu]       (yaml)  {YAML};
  \node[nd,  right=of yaml]     (parse) {parse};
  \node[nd,  right=of parse]    (valid) {validate};
  \node[nd,  right=of valid]    (comp)  {compile};
  \node[cdn, below=17mm of hu]  (ir)    {JSON IR};
  \node[nd,  right=of ir]       (eng)   {Engine};
  \node[nd,  right=30mm of eng] (ad)    {Adapter};
  \node[nd,  right=of ad]       (ai)    {AI-system};
  \draw[bi] (hu)    -- (yaml);
  \draw[fl] (yaml)  -- (parse);
  \draw[fl] (parse) -- (valid);
  \draw[fl] (valid) -- (comp);
  \draw[fl] (comp.south) -- ++(0,-8mm) -| (ir.north);
  \draw[fl] (ir)  -- (eng);
  \draw[fl] (eng) -- node[fill=white, inner sep=1.5pt, font=\scriptsize] {runtime contract} (ad);
  \draw[fl] (ad)  -- (ai);
\end{tikzpicture}
\end{center}

The \textbf{engine} is Benchy's execution machinery. It consumes the compiled JSON IR, reads exam examples, invokes the AI-system through an adapter, validates outputs, computes scores, and produces results.

The \textbf{runtime contract} is the interface rule the engine expects at its boundary:

\begin{equation}
\boxed{\text{named-field input object}\rightarrow\text{named-field output object}}
\end{equation}
\gloss{Every Benchy-compliant AI-system is presented to the engine as something that consumes an object matching the program's named input fields and produces an object matching its named output fields.}

An adapter translates between that contract and an AI-system's native interface.

\section{Semantic object model}

The program defines what must be done.

\begin{equation}
\boxed{\text{Program}=\text{Input Schema}+\text{Output Schema}}
\end{equation}
\gloss{A program consists of one schema describing its inputs and another describing its outputs.}

Conceptually:

\begin{equation}
P:X\rightarrow Y
\end{equation}
\gloss{Program $P$ maps values conforming to input schema $X$ into values conforming to output schema $Y$.}

A schema combines field structure with semantic types:

\begin{equation}
\boxed{\text{Schema}=\text{Field Structure}+\text{Semantic Types}}
\end{equation}
\gloss{A schema says which named fields exist, how they are nested, and what each field semantically means.}

The semantic type vocabulary is:

\begin{center}
\begin{tabular}{ll}
\toprule
\textbf{Type} & \textbf{Meaning} \\
\midrule
\texttt{string} & unconstrained text \\
\texttt{int} & integer \\
\texttt{float} & real-valued number \\
\texttt{bool} & true / false \\
\texttt{enum} & value from a closed categorical set \\
\texttt{date} & calendar date \\
\texttt{time} & time of day \\
\texttt{datetime} & date + time \\
\texttt{image} & one visual artifact treated as an image \\
\texttt{audio} & audio artifact \\
\texttt{document} & document artifact that may contain pages, text, images, and layout \\
\bottomrule
\end{tabular}
\end{center}

For an enum:

\begin{equation}
Y=\{v_1,v_2,\ldots,v_k\}
\end{equation}
\gloss{An enum field may take exactly one value from the finite declared set.}

Every input and output value is represented through one or more named fields.

\begin{verbatim}
program:
  input:
    audio: audio
  output:
    transcription: string
\end{verbatim}

A classification program can be:

\begin{verbatim}
program:
  input:
    text: string
  output:
    sentiment:
      enum: [positive, neutral, negative]
\end{verbatim}

A structured extraction program can be:

\begin{verbatim}
program:
  input:
    image: image
  output:
    invoice_number: string
    date: date
    supplier: string
    subtotal: float
    total: float
\end{verbatim}

Fixed nested structures are allowed:

\begin{verbatim}
program:
  input:
    document: document
  output:
    supplier:
      name: string
      tax_id: string
    date: date
    total: float
\end{verbatim}

A \textbf{leaf output field} is an output field whose value is a semantic type rather than another nested field structure.

The leaf output fields are the scoring dimensions.

\begin{equation}
\boxed{\text{fixed named output schema}\Rightarrow\text{fixed scoring dimensions}\Rightarrow\text{fixed field weights}}
\end{equation}
\gloss{Once the output fields are fixed, Benchy knows exactly which dimensions can be scored and weighted.}

\section{Benchmark classification and the shared SURUS ontology}

Benchmarks are classified with the shared SURUS task-oriented AI ontology:

\begin{equation}
\boxed{/\text{task}/\text{domain}/\text{language}/}
\end{equation}
\gloss{The ontology identifies what operation is performed, the domain/distribution in which it is evaluated, and the linguistic context of the data.}

\begin{verbatim}
benchmark:
  task: extract
  domain: finance
  language: es
\end{verbatim}

The three coordinates have different roles:

\begin{itemize}
  \item \textbf{task} describes the operation performed;
  \item \textbf{domain} describes the world or distribution represented by the exam data;
  \item \textbf{language} describes the linguistic distribution represented by the exam data.
\end{itemize}

Tasks, domains, and languages live in one shared, versioned registry used by Benchy, DataHub, EvalsHub, and other SURUS systems.

A task defines a family of admissible programs.

Let:

\begin{itemize}
  \item $T$ = a task;
  \item $\mathcal{P}_T$ = the family of programs admitted by task $T$;
  \item $P$ = a concrete program.
\end{itemize}

Then:

\begin{equation}
T\rightarrow\mathcal{P}_T
\end{equation}
\gloss{Task $T$ determines the family of programs considered valid instances of that task.}

The compiler checks:

\begin{equation}
\boxed{P\in\mathcal{P}_T}
\end{equation}
\gloss{Program $P$ must belong to the family admitted by task $T$.}

The program schema alone cannot always determine the task.

\begin{equation}
\text{string}\rightarrow\text{string}
\end{equation}
\gloss{The same input/output shape could represent translation, summarization, rewriting, question answering, or another operation.}

Task is therefore explicit rather than inferred.

The ontology registry carries shared identifiers and descriptions. Benchy implements the finite task-to-program structural validation rules for the ontology versions it supports. The registry does not contain a separate constraint language.

Translation uses an ordered source/target language relation:

\begin{verbatim}
benchmark:
  task: translate
  domain: general
  language:
    source: es
    target: en
\end{verbatim}

which may be represented externally as:

\begin{verbatim}
/translate/general/es-en
\end{verbatim}

\section{Scoring}

The output schema determines the dimensions over which correctness can be evaluated.

A scoring function specifies the relative importance of output fields and the rule by which field-level correctness is aggregated into program-level performance.

\begin{equation}
\boxed{\text{Scoring Function}=\text{Field Weights}+\text{Aggregator}}
\end{equation}
\gloss{Field weights express how important each output dimension is; the aggregator defines how those field scores become one instance score.}

\begin{verbatim}
scoring:
  weights:
    invoice_number: 1
    date: 1
    supplier: 1
    subtotal: 1
    total: 5
  aggregator: weighted_mean
\end{verbatim}

Nested weights mirror the output schema:

\Needspace*{14\baselineskip}
\begin{verbatim}
program:
  output:
    supplier:
      name: string
      tax_id: string
    total: float

scoring:
  weights:
    supplier:
      name: 1
      tax_id: 0
    total: 5
  aggregator: weighted_mean
\end{verbatim}

Every leaf output field has exactly one explicit weight. There are no missing weights, extra weights, or weights on intermediate objects.

Weights are non-negative:

\begin{equation}
w_j\geq0
\end{equation}
\gloss{The weight $w_j$ of scoring dimension $j$ may be zero or positive, but never negative.}

A zero weight means the field is present and validated but does not affect the instance score.

For exam example $i$ and scoring dimension $j$, field correctness is:

\begin{equation}
c_{ij}=\mathbf{1}[\hat y_{ij}=y_{ij}^*]
\end{equation}
\gloss{Field correctness $c_{ij}$ is 1 when the predicted semantic value exactly equals the expected semantic value and 0 otherwise.}

Field correctness is currently produced by exact match. The evaluator is not exposed as a benchmark configuration option.

All field scores for example $i$ form:

\begin{equation}
\mathbf{c}_i=(c_{i1},c_{i2},\ldots,c_{in})
\end{equation}
\gloss{$\mathbf{c}_i$ contains one correctness value for each scored output dimension of exam example $i$.}

The current aggregator is normalized weighted mean:

\begin{equation}
s_i=\frac{\sum_j w_jc_{ij}}{\sum_jw_j}
\end{equation}
\gloss{Instance score $s_i$ is the weighted average of field correctness values for exam example $i$.}

Therefore:

\begin{equation}
\sum_jw_j>0
\end{equation}
\gloss{At least one scoring dimension must have positive weight.}

The weights encode \textbf{relative importance}. Multiplying every weight by the same positive constant does not change the instance score.

Normalization is a property of this aggregator, not a universal property of Benchy.

\begin{equation}
\boxed{\text{The aggregator defines the semantics and scale of the score.}}
\end{equation}
\gloss{A normalized aggregator can produce a score in $[0,1]$; future aggregators may naturally use another scale.}

\section{Data}

The dataset is the exam.

Each exam example $i$ contains:

\begin{equation}
(x_i,y_i^*)
\end{equation}
\gloss{$x_i$ is the input for example $i$, and $y_i^*$ is its expected or ground-truth output.}

The program schema constrains both sides:

\begin{equation}
x_i\in X
\end{equation}
\gloss{Exam input $x_i$ must conform to the program's input schema $X$.}

\begin{equation}
y_i^*\in Y
\end{equation}
\gloss{Expected output $y_i^*$ must conform to the program's output schema $Y$.}

Thus:

\begin{verbatim}
Input Schema  ---> exam inputs
Output Schema ---> expected outputs
Output Schema ---> scoring dimensions
\end{verbatim}

The benchmark definition references the dataset:

\begin{verbatim}
data:
  path: ./data/invoices.jsonl
\end{verbatim}

The program schema defines the semantic structure of every example. The engine specification in Appendix~\ref{app:engine} defines the concrete dataset encoding used by the current implementation.

\section{AI-system}

The AI-system is the benchmark taker.

\begin{equation}
\mathrm{AI}:X\rightarrow Y
\end{equation}
\gloss{AI-system $\mathrm{AI}$ implements the declared program contract: it consumes values from $X$ and produces values intended to conform to $Y$.}

An AI-system may be:

\begin{verbatim}
single AI model
AI-node: model + optimized prompt + fixed task behavior
composition of AI models or AI programs
AI components + explicit deterministic code
agent or workflow
\end{verbatim}

For a directly specified model:

\begin{verbatim}
ai-system:
  type: model
  provider: openai
  model: <model>
  prompt: ./prompts/invoice.md
  parameters:
    temperature: 0
\end{verbatim}

These fields identify the AI-system under evaluation.

Credentials, SDK construction, HTTP mechanics, response extraction, and similar integration details are outside benchmark semantics.

An arbitrary external AI-system can instead be identified as:

\begin{verbatim}
ai-system:
  type: external
  id: invoice-extractor-v7
\end{verbatim}

The runtime environment binds that identifier to an adapter.

The distinction is:

\begin{verbatim}
AI-system definition
= what is being evaluated

adapter/runtime binding
= how this environment invokes it
\end{verbatim}

\section{Compilation and execution architecture}

The canonical compilation path is:

\begin{equation}
\boxed{\text{YAML}\rightarrow\text{parse}\rightarrow\text{validate}\rightarrow\text{compile}\rightarrow\text{JSON IR}\rightarrow\text{engine}}
\end{equation}
\gloss{Benchy parses the canonical source, validates its semantics, compiles it once into a machine-facing representation, and executes from that representation.}

There is no normalization stage.

Let:

\begin{itemize}
  \item $Y_{\text{yaml}}$ = valid Benchy YAML;
  \item $C$ = compiler;
  \item $J$ = canonical JSON IR.
\end{itemize}

Then:

\begin{equation}
\boxed{J=C(Y_{\text{yaml}})}
\end{equation}
\gloss{Compiler $C$ deterministically produces JSON IR $J$ from valid YAML $Y_{\text{yaml}}$.}

Compilation preserves meaning:

\begin{equation}
\boxed{\operatorname{Semantics}(J)=\operatorname{Semantics}(Y_{\text{yaml}})}
\end{equation}
\gloss{JSON IR and YAML express the same benchmark/run semantics; the IR is the executable representation, not another semantic source.}

The engine does not reinterpret YAML.

At runtime, the engine knows one AI-system interface:

\begin{equation}
\boxed{\text{named-field input object}\rightarrow\text{named-field output object}}
\end{equation}
\gloss{External AI-system differences are translated at the adapter boundary instead of being spread through the engine.}

\section{Execution and results}

For exam example $i$:

\begin{equation}
\hat y_i=\mathrm{AI}(x_i)
\end{equation}
\gloss{AI-system $\mathrm{AI}$ receives input $x_i$ and produces prediction $\hat y_i$.}

The output is validated against schema $Y$ before scoring.

A valid but completely incorrect output can have:

\begin{verbatim}
status: valid
score: 0
\end{verbatim}

An output that does not satisfy the declared schema is structurally different:

\begin{verbatim}
status: invalid_output
score: null
\end{verbatim}

An invocation failure is different again:

\begin{verbatim}
status: execution_error
score: null
\end{verbatim}

This distinction preserves the difference between a valid zero score and a failure to produce a valid program output.

For final aggregation, define each example's contribution $q_i$:

\begin{equation}
q_i=
\begin{cases}
s_i, & \text{if the AI-system produced a valid output} \\
0, & \text{if the output was invalid or execution failed}
\end{cases}
\end{equation}
\gloss{Invalid outputs and execution failures retain \texttt{null} as their stored instance score, but contribute zero to the benchmark aggregate.}

For $N$ exam examples:

\begin{equation}
\boxed{B(\mathrm{AI})=\frac{1}{N}\sum_{i=1}^{N}q_i}
\end{equation}
\gloss{The benchmark score is the arithmetic mean across all example contributions, so failed examples remain in the denominator.}

\Needspace*{36\baselineskip}
\section{Canonical YAML}

\begin{verbatim}
version: "1.0"
ontology_version: "1.0"

benchmark:
  task: extract
  domain: finance
  language: es

program:
  input:
    image: image
  output:
    invoice_number: string
    date: date
    supplier: string
    subtotal: float
    total: float

scoring:
  weights:
    invoice_number: 1
    date: 1
    supplier: 1
    subtotal: 1
    total: 5
  aggregator: weighted_mean

data:
  path: ./data/invoices.jsonl

ai-system:
  type: external
  id: invoice-extractor-v7
\end{verbatim}

\texttt{version} pins the Benchy semantic specification.

\texttt{ontology\_version} pins the shared SURUS task/domain/language registry used to interpret the benchmark classification.

\section{Scope and future extensions}

The current language deliberately does not expose:

\begin{verbatim}
variable-length output collections
custom field evaluators
aggregators other than weighted_mean
adapter configuration inside benchmark YAML
implicit field weights
anonymous root scalar inputs or outputs
\end{verbatim}

These are extension points rather than undefined behavior.

The present architecture is intended to let those capabilities be introduced without changing the meaning of existing benchmark definitions.

\appendix

\section{Engine 1.0 engineering contract}
\label{app:engine}

This appendix is normative for the first implementation. The main paper defines semantics; this appendix fixes the engineering representation required to build the engine.

\subsection{YAML grammar and strictness}

The top-level YAML object contains exactly:

\begin{verbatim}
version
ontology_version
benchmark
program
scoring
data
ai-system
\end{verbatim}

Unknown top-level keys are errors.

Mappings must not contain duplicate keys.

YAML anchors, aliases, merge keys, and custom tags are rejected.

The program grammar is:

\begin{verbatim}
program:
  input:
    <field-name>: <type-or-nested-object>
  output:
    <field-name>: <type-or-nested-object>
\end{verbatim}

Rules:

\begin{itemize}
  \item \texttt{input} and \texttt{output} are mappings;
  \item each contains at least one named field;
  \item field names are non-empty strings;
  \item all fields are required;
  \item \texttt{null} is not a valid field value;
  \item lists/arrays are not supported as field values;
  \item nested objects are allowed;
  \item scalar type declarations must be one of the supported semantic type names;
  \item enum is declared only as:
\end{itemize}

\begin{verbatim}
sentiment:
  enum: [positive, neutral, negative]
\end{verbatim}

Enum values are non-empty, distinct strings.

Unknown type declarations are errors.

\subsection{Strict object validation}

Program schemas are closed.

For dataset records and AI-system outputs:

\begin{verbatim}
missing required field -> invalid
extra field            -> invalid
wrong semantic type    -> invalid
\end{verbatim}

Nested objects are validated recursively.

\subsection{Weight coverage}

The weight tree must mirror the output schema down to every leaf output field.

There must be:

\begin{verbatim}
exactly one explicit weight per leaf output field
no missing weights
no extra weights
no weights on intermediate objects
\end{verbatim}

Weights are finite numbers satisfying $w_j\ge0$, and their total must be positive.

\subsection{Task-to-program validation}

The shared ontology registry contains identifiers and descriptions only.

Benchy implements the structural validators for the task vocabulary of the ontology version it supports.

For ontology version \texttt{1.0}:

\begin{verbatim}
extract
  no task-specific structural constraint beyond normal program rules

classify
  exactly one leaf output field
  that leaf field is enum

transcribe
  input contains at least one audio leaf field
  exactly one leaf output field
  that leaf field is string

translate
  input contains at least one string leaf field
  exactly one leaf output field
  that leaf field is string
  benchmark.language is {source, target}
\end{verbatim}

If an ontology task is not supported by the loaded Benchy specification, compilation fails rather than silently skipping task validation.

\subsection{Dataset encoding}

The engine uses JSONL.

Each non-empty line is one exam example:

\begin{verbatim}
{
  "input": {
    "image": "assets/invoice-001.png"
  },
  "expected": {
    "total": 121.0
  }
}
\end{verbatim}

The row contains exactly \texttt{input} and \texttt{expected}.

Blank lines may be ignored.

The dataset must contain at least one example.

Dataset examples are streamed. They are not embedded into JSON IR.

Each example is validated at execution time against the already-compiled schemas in the IR.

An invalid dataset record is a benchmark/data error and aborts the run. It is not scored as an AI-system failure.

\subsection{JSON representation of semantic values}

\begin{center}
\begin{tabular}{ll}
\toprule
\textbf{Semantic type} & \textbf{JSON/runtime representation} \\
\midrule
\texttt{string} & string \\
\texttt{enum} & string \\
\texttt{int} & integer \\
\texttt{float} & finite JSON number; integers are valid real values \\
\texttt{bool} & boolean \\
\texttt{date} & canonical string \texttt{YYYY-MM-DD} \\
\texttt{time} & canonical string \texttt{HH:MM:SS[.fraction]} \\
\texttt{datetime} & RFC 3339 string with timezone \\
nested object & JSON object \\
\texttt{image} & filesystem path string \\
\texttt{audio} & filesystem path string \\
\texttt{document} & filesystem path string \\
\bottomrule
\end{tabular}
\end{center}

\texttt{bool} is not accepted as \texttt{int} or \texttt{float}.

\subsubsection*{Artifact path rule}

In a dataset, \texttt{image}, \texttt{audio}, and \texttt{document} are encoded as relative path strings.

The engine:

\begin{enumerate}
  \item resolves the path relative to the JSONL file's directory;
  \item normalizes it;
  \item verifies that it exists and is a regular readable file;
  \item passes the resolved absolute path string through the runtime contract.
\end{enumerate}

The adapter decides whether to open the file, upload it, decode it, or convert it to another native representation.

For artifact outputs, an adapter returns a local path string to the produced file. The engine validates that the file exists before treating the output as schema-valid.

This is an implementation representation of the semantic artifact types, not a redefinition of those types.

\subsection{Exact-match equality}

Exact match compares schema-valid semantic values.

\begin{center}
\begin{tabular}{ll}
\toprule
\textbf{Type} & \textbf{Equality} \\
\midrule
\texttt{string} & exact Unicode string equality \\
\texttt{enum} & exact declared-member equality \\
\texttt{int} & integer equality \\
\texttt{float} & finite numeric equality \\
\texttt{bool} & boolean equality \\
\texttt{date} & parsed calendar-date equality \\
\texttt{time} & parsed time-of-day equality \\
\texttt{datetime} & parsed instant equality \\
\texttt{image} & byte-for-byte file equality \\
\texttt{audio} & byte-for-byte file equality \\
\texttt{document} & byte-for-byte file equality \\
\bottomrule
\end{tabular}
\end{center}

There is no trimming, case folding, Unicode normalization, numeric tolerance, or semantic similarity.

\subsection{Compile-time versus execution-time validation}

Compile-time validation covers:

\begin{verbatim}
YAML syntax
specification version
ontology version
task/domain/language registry membership
task <-> program compatibility
program schema grammar
weight coverage and validity
data configuration
AI-system semantic definition
\end{verbatim}

Execution-time validation covers:

\begin{verbatim}
each dataset input against compiled input schema
each expected value against compiled output schema
each AI-system output against compiled output schema
\end{verbatim}

The engine uses schemas already present in the JSON IR. It never reinterprets source YAML.

\subsection{Canonical JSON IR}

The IR contains the validated executable semantics needed by the engine, including:

\begin{verbatim}
versions
benchmark classification
typed input/output schemas
resolved scoring dimensions
field evaluator
instance aggregator
benchmark aggregator
data configuration
AI-system semantic definition
\end{verbatim}

Nested output fields become scoring paths represented as string arrays, for example:

\begin{verbatim}
{
  "path": ["supplier", "name"],
  "weight": 1
}
\end{verbatim}

Paths are an IR representation detail. Authors do not write them.

The IR includes:

\Needspace*{10\baselineskip}
\begin{verbatim}
{
  "scoring": {
    "evaluator": "exact_match",
    "instance_aggregator": "weighted_mean",
    "benchmark_aggregator": "mean"
  },
  "data": {
    "format": "jsonl"
  }
}
\end{verbatim}

These fields make fixed specification semantics explicit to the engine; they do not introduce new benchmark meaning.

\subsection{Adapter protocol}

The engine binds one adapter before execution.

Conceptually:

\begin{verbatim}
class Adapter:
    async def invoke(
        self,
        input_object: dict[str, object],
    ) -> dict[str, object]:
        ...
\end{verbatim}

The adapter instance may be created with the compiled program contract and AI-system definition.

The engine owns:

\begin{verbatim}
dataset loading
input validation
output validation
scoring
aggregation
result/error recording
\end{verbatim}

The adapter owns:

\begin{verbatim}
mapping Benchy inputs to the native AI-system interface
invoking the AI-system
mapping native outputs back to the Benchy named-field object
\end{verbatim}

Adapter exceptions become execution errors.

If the adapter returns an object that violates the output schema, the result is \texttt{invalid\_output}.

Credentials, endpoint mappings, response paths, Python imports, SDK construction, and other integration mechanics are runtime configuration outside benchmark YAML.

The default engine may execute sequentially. Concurrency is an execution optimization and must not change benchmark semantics.

\subsection{AI-system execution setup}

For:

\begin{verbatim}
ai-system:
  type: external
  id: invoice-extractor-v7
\end{verbatim}

the runtime must provide an adapter binding for that identifier before the run begins.

Missing binding is a run setup error and aborts before evaluating examples.

For:

\begin{verbatim}
ai-system:
  type: model
  provider: ...
  model: ...
\end{verbatim}

a provider adapter may be selected by the runtime. Provider integrations are built on the same adapter protocol and are outside the core engine.

\subsection{Instance statuses}

Every evaluated example has one of:

\begin{center}
\texttt{valid}\quad \texttt{invalid\_output}\quad \texttt{execution\_error}
\end{center}

\texttt{valid} means the AI-system produced a schema-valid output, regardless of score.

\texttt{invalid\_output} means the adapter returned an output value but it failed the compiled output schema.

\texttt{execution\_error} means invocation failed before a Benchy output object was successfully produced.

Examples include adapter exceptions, HTTP/provider failures, and runtime timeouts.

\subsection{Scoring failed examples}

Stored instance score:

\begin{verbatim}
valid            -> numeric s_i
invalid_output   -> null
execution_error  -> null
\end{verbatim}

Aggregation contribution:

\begin{equation}
q_i=
\begin{cases}
s_i, & \text{valid} \\
0, & \text{invalid\_output or execution\_error}
\end{cases}
\end{equation}
\gloss{\texttt{null} preserves the diagnostic distinction, while zero contribution prevents failed examples from disappearing from the aggregate.}

\Needspace*{16\baselineskip}
\subsection{Benchmark result schema}

A run result contains:

\begin{verbatim}
{
  "version": "1.0",
  "benchmark_score": 0.86,
  "summary": {
    "examples": 100,
    "valid": 96,
    "invalid_outputs": 2,
    "execution_errors": 2
  },
  "results": []
}
\end{verbatim}

A valid example result:

\Needspace*{16\baselineskip}
\begin{verbatim}
{
  "index": 0,
  "status": "valid",
  "prediction": {
    "total": 100.0
  },
  "field_scores": [
    {
      "path": ["total"],
      "score": 0,
      "weight": 5
    }
  ],
  "score": 0.0,
  "contribution": 0.0,
  "error": null
}
\end{verbatim}

An invalid output:

\Needspace*{14\baselineskip}
\begin{verbatim}
{
  "index": 0,
  "status": "invalid_output",
  "prediction": {
    "debug": "foo"
  },
  "field_scores": null,
  "score": null,
  "contribution": 0.0,
  "error": {
    "code": "missing_field",
    "path": ["total"],
    "message": "required output field is missing"
  }
}
\end{verbatim}

An execution error:

\Needspace*{13\baselineskip}
\begin{verbatim}
{
  "index": 0,
  "status": "execution_error",
  "prediction": null,
  "field_scores": null,
  "score": null,
  "contribution": 0.0,
  "error": {
    "code": "adapter_error",
    "path": null,
    "message": "..."
  }
}
\end{verbatim}

\subsection{Dataset failures}

Dataset failures are run-level benchmark errors rather than example scores.

Examples:

\begin{verbatim}
data file missing
malformed JSONL
row lacks input/expected
input does not conform to input schema
expected output does not conform to output schema
artifact path does not exist
empty dataset
\end{verbatim}

The run aborts and does not produce a benchmark score.

\subsection{Error model}

Compiler and runtime diagnostics should use:

\begin{center}
\texttt{phase}\quad \texttt{code}\quad \texttt{path}\quad \texttt{message}
\end{center}

\texttt{path} is an array of semantic/object keys when applicable.

Representative codes:

\medskip
\noindent{\small\ttfamily\raggedright\leftskip=1.5em
invalid\_yaml\quad duplicate\_key\quad unknown\_key\quad unsupported\_version\quad unsupported\_ontology\_version\quad unknown\_task\quad unknown\_domain\quad unknown\_language\quad task\_program\_mismatch\quad invalid\_schema\quad missing\_weight\quad extra\_weight\quad invalid\_weight\quad data\_not\_found\quad invalid\_dataset\_record\quad artifact\_not\_found\quad missing\_field\quad extra\_field\quad wrong\_type\quad invalid\_enum\quad invalid\_value\quad adapter\_error\quad timeout\quad provider\_error\par}
\medskip

The exact message text is not part of benchmark semantics.

\section{Shared SURUS ontology registry}

The ontology is one shared YAML file containing:

\begin{center}
\texttt{tasks}\quad \texttt{domains}\quad \texttt{languages}
\end{center}

Example:

\Needspace*{22\baselineskip}
\begin{verbatim}
version: "1.0"

tasks:
  extract:
    description: Extract named information from an input.
  classify:
    description: Assign one declared categorical label.
  transcribe:
    description: Convert spoken audio into text.
  translate:
    description: Transform text from a source language to a target language.

domains:
  general: {}
  finance: {}
  healthcare: {}
  legal: {}
  retail: {}

languages:
  es: {}
  pt: {}
  en: {}
\end{verbatim}

Benchy code, not the registry, implements the structural task validators for each supported ontology version.

This avoids prematurely introducing a task-constraint DSL while preserving:

\begin{equation}
P\in\mathcal{P}_T
\end{equation}
\gloss{The declared program must satisfy the structural semantics associated with its task.}

\section{Artifact runtime representation}

Two designs were considered for \texttt{image}, \texttt{audio}, and \texttt{document}.

\subsection*{Typed artifact object}

The engine could convert a dataset path into an internal object such as:

\begin{verbatim}
Artifact(type=image, path=...)
\end{verbatim}

This gives the runtime value an explicit host-language wrapper, but introduces a new abstraction that every adapter and language binding must understand.

\subsection*{Path string}

The dataset may keep the artifact as a path string and let the adapter decide how to consume it.

This is simpler and fits the adapter boundary: one adapter may open the file, another may upload it over HTTP, and another may decode it into a provider-specific object.

The first implementation therefore uses \textbf{resolved path strings} at the runtime boundary.

The semantic meaning still comes from the program schema:

\begin{verbatim}
image: image
\end{verbatim}

not from the host-language type of the value.

A future runtime representation may replace path strings with a richer artifact handle without changing the benchmark's semantic type system.

\section{Compact architecture}

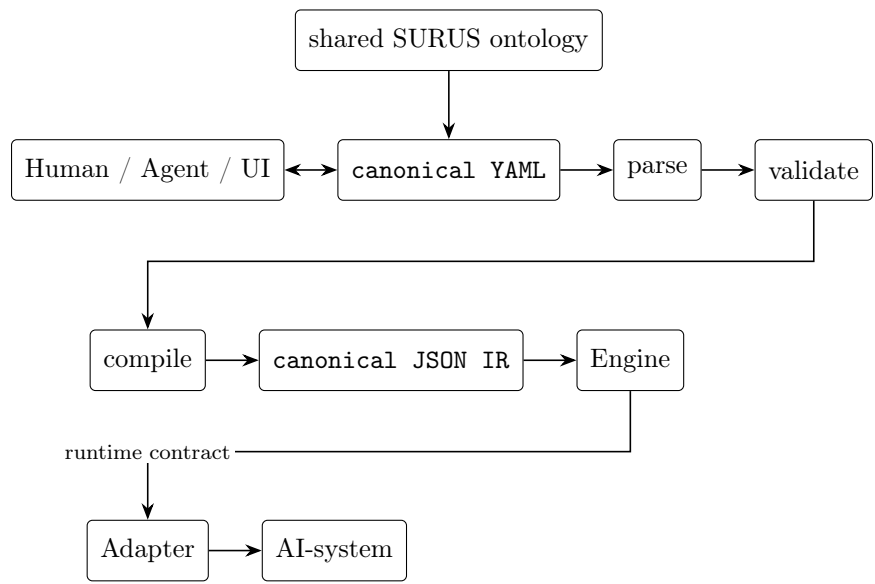
\begin{center}
\begin{tikzpicture}[node distance=6mm and 7mm]
  \node[nd]                      (ont)   {shared SURUS ontology};
  \node[cdn, below=9mm of ont]   (yaml)  {canonical YAML};
  \node[nd,  left=of yaml]       (hu)    {Human / Agent / UI};
  \node[nd,  right=of yaml]      (parse) {parse};
  \node[nd,  right=of parse]     (valid) {validate};
  \node[nd,  below=17mm of hu]   (comp)  {compile};
  \node[cdn, right=of comp]      (ir)    {canonical JSON IR};
  \node[nd,  right=of ir]        (eng)   {Engine};
  \node[nd,  below=17mm of comp] (ad)    {Adapter};
  \node[nd,  right=of ad]        (ai)    {AI-system};
  \draw[fl] (ont)   -- (yaml);
  \draw[bi] (hu)    -- (yaml);
  \draw[fl] (yaml)  -- (parse);
  \draw[fl] (parse) -- (valid);
  \draw[fl] (valid.south) -- ++(0,-8mm) -| (comp.north);
  \draw[fl] (comp)  -- (ir);
  \draw[fl] (ir)    -- (eng);
  \draw[fl] (eng.south) -- ++(0,-8mm)
        -| node[pos=0.5, fill=white, inner sep=1.5pt, font=\scriptsize] {runtime contract} (ad.north);
  \draw[fl] (ad)    -- (ai);
\end{tikzpicture}
\end{center}

The central rule is:

\gloss{\textbf{Benchmark meaning is defined once in the canonical YAML, compiled once into JSON IR, and executed through one runtime contract.}}

\end{document}